\documentclass[runningheads]{llncs}

\usepackage{eccv}

\usepackage{eccvabbrv}

\usepackage{graphicx}
\usepackage{booktabs}
\usepackage{multirow}
\usepackage{makecell}
\usepackage{comment}
\usepackage{placeins}
\usepackage[accsupp]{axessibility}  % Improves PDF readability for those with disabilities.

\newcommand{\steph}[1]{\textcolor{DarkOrchid}{[Steph]: #1}}

\usepackage{hyperref}

\usepackage{orcidlink}

\begin{document}

\title{Reasoning for Social Audio-Visual Question Answering: Where Do We Stand?} 

\titlerunning{Reasoning for Social AV-QA: Where Do We Stand?}

\author{Koen P. de Vries\inst{1,2} \and
Xavier Alameda-Pineda \inst{1} \and
Estefanía Talavera \inst{2} \and
Stéphane Lathuilière \inst{1}}

\authorrunning{K. P. de Vries et al.}

\institute{Inria at Univ. Grenoble Alpes, CNRS, LJK, France \and University of Twente, The Netherlands 
}

\maketitle
%\steph{the title does not capture the intentbench section. }
%\steph{idea: 
%\begin{enumerate}
%    \item HumanOmni simpler, faster, cleaner 
%    \item Frustratingly easy social audio-visual understanding. 
%    \item Where are we at social audio-visual reasoning and understanding
%    \item A Reality Check on Social Audio-Visual Reasoning and Understanding 
%    \item Social Audio-Visual Reasoning: Where Do We Stand? 
%    \end{enumerate}
%}
\newcommand{\benchname}{\textsc{IntentBench-Prime}\xspace}
\newcommand{\intentbench}{\textsc{IntentBench}\xspace}
\newcommand{\worldsense}{\textsc{WorldSense}\xspace}
\newcommand{\dailyomni}{\textsc{Daily-Omni}\xspace}

\begin{abstract}
  Training Multimodal Large Language Models for audio-visual social understanding is a crucial step toward embodied social intelligence. Chain-of-thought (CoT) reasoning has become the dominant approach, with HumanOmniV2 and its IntentBench benchmark as a prominent reference point. In this context, we report three findings. First, IntentBench is highly noisy: $\sim$7\% of questions are broken and $\sim$23\% are trivially answerable without the video input. We remove the affected questions and release \benchname. Second, current reasoning approaches are expensive and surprisingly ineffective. A simple \emph{Vanilla SFT} baseline matches or outperforms existing reasoning methods across three benchmarks at a fraction of the cost, establishing it as an essential baseline for evaluating novel fine-tuning techniques. Third, our analysis reveals that substantial priors can be learned solely from the text modality and that using a textual caption instead of the video yields performance on par with Vanilla SFT. These surprising findings reveal the limitations of current MLLMs when it comes to social understanding. \href{https://github.com/koenv759/IntentBench-Prime}{IntentBench-Prime}, \href{https://huggingface.co/koenv759/VanillaSFT-LoRA}{\emph{Vanilla SFT} model}, and \href{https://github.com/koenv759/VanillaSFT}{code} are publicly available.

\end{abstract}

\begin{figure}[h!]
  \centering
  \includegraphics[width=\textwidth]{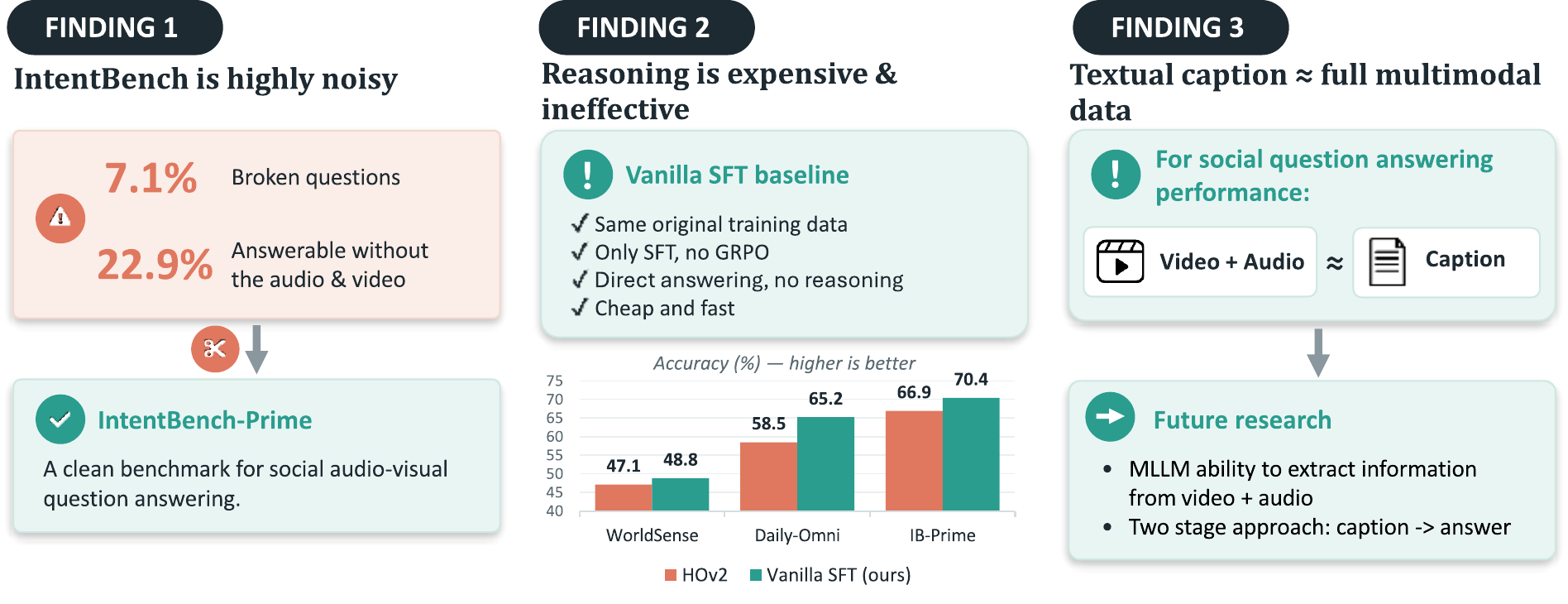}
  \vspace{-.3in}  
  \caption{Social \textit{Audio-Visual Question Answering} (AV-QA): Our three key findings.}
  \label{fig: teaser}
\end{figure}

\section{Introduction}
\label{sec: intro}
Understanding human intentions, emotions, and complex social scenarios is a crucial step toward embodied social intelligence. Multimodal Large Language Models (MLLMs), or Omni-models, have emerged as highly effective tools for this objective. They can natively process video, audio, and textual inputs to generate text or even audio, as demonstrated by the Qwen-omni series \cite{qwen2025qwen25omni}. %Concurrently, the community is increasingly focused on developing strong reasoning capabilities. Inspired by DeepSeek-R1 \cite{guo2025deepseek}, which successfully scaled general reasoning training via Group Relative Policy Optimization (GRPO) \cite{shao2024deepseekmathpushinglimitsmathematical}, recent works have begun translating these techniques to the multimodal domain \cite{feng2026videor1,yang2025humanomniv2,luo2026unveiling,xiao2025perceptionr1} to enhance the analytical capacities of MLLMs. 
\\ \\
In this context, several works focus specifically on multimodal social understanding \cite{yang2025humanomniv2, 11585891, li2025vegas, li2025towards} to improve how MLLMs respond to human-centric queries. 
Notably, HumanOmniV2 \cite{yang2025humanomniv2} has significantly impacted this field \cite{11585891,ma2026modf,Rha_Yeo_Kim_Ro_2026,kulkarni2026avatar} due to its strong performance and the introduction of \intentbench, a benchmark specifically aimed at evaluating multimodal social understanding capabilities.
%Notably, HumanOmniV2 \cite{yang2025humanomniv2} has significantly impacted this field \cite{X,Y,Z} due to its strong performance and extensive evaluation efforts. To benchmark fine-grained social intelligence, HumanOmniV2 introduced \intentbench, a dataset rapidly adopted by subsequent research \cite{X,Y,Z}. 
However, our in-depth analysis of this benchmark reveals a critical limitation, leading to our first key finding (\cref{fig: teaser}): %\steph{we could take the Fining 1 black logo from the teaser}
\raisebox{-0.28\height}{\includegraphics[height=1.1\baselineskip]{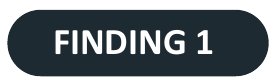}}~\textbf{\intentbench evaluation is highly noisy.} We identify several issues within \intentbench, specifically broken or poorly formulated questions ($\sim$7\%) originating both from the Social-IQ 2.0 source data and the curation process. Furthermore, we show that for a substantial portion of the benchmark ($\sim$23\%), pretrained LLMs unanimously agree on the correct answer without any video or audio inputs, relying entirely on language priors. Motivated by these observations, we introduce a rigorously curated version of this dataset, denoted as \benchname. \\ \\
Methodologically, recent approaches \cite{yang2025humanomniv2,qin2026humansense, li2025vegas, 11585891, luo2026unveiling} have integrated reasoning techniques into MLLMs to enhance social Audio-Visual Question Answering (AV-QA). These models are typically trained to perform reasoning via a Chain-of-Thought \cite{3600270.3602070} (CoT) strategy optimized with GRPO \cite{guo2025deepseek, shao2024deepseekmathpushinglimitsmathematical}. In the case of HumanOmniV2 \cite{yang2025humanomniv2}, the model is tasked to first summarize relevant context before performing reasoning.  While intended to mitigate insufficient context comprehension and shortcut reasoning, CoT introduces substantial overhead. In this work, we explore an alternative direction: improving social AV-QA without the computational burden of CoT. Our experiments yield a second key finding: \raisebox{-0.28\height}{\includegraphics[height=1.1\baselineskip]{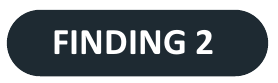}}~\textbf{Current CoT-based reasoning for social understanding is expensive and ineffective.} To demonstrate this, we introduce a frustratingly straightforward yet successful baseline, termed \emph{Vanilla SFT}. This baseline is trained using standard supervised fine-tuning (SFT) for direct AV-QA, leveraging the exact same AV-QA data as HumanOmniV2 but without reasoning traces. Through rigorous evaluation across three benchmarks, \benchname,  \worldsense \cite{hong2026worldsense}, and \dailyomni \cite{zhou2025daily}, we show that \emph{Vanilla SFT} matches or outperforms existing reasoning approaches. Crucially, it does so while being vastly more efficient. It circumvents the need to distill reasoning traces, accelerates training time, and dramatically reduces inference latency by bypassing the auto-regressive generation of lengthy rationales. Based on these results, we strongly advocate for the establishment of \emph{Vanilla SFT} as an essential baseline when evaluating novel fine-tuning methods. Despite its simplicity, such grounding is key for measuring methodological progress in this domain. \\ %\koen{The finding also seems to hold for non-social stuff (WorldSense and Daily-omni), add something on that? | Added the word 'current' in front of the finding} 
Finally, motivated by Finding 2 and aiming to understand more thoroughly to what extent MLLMs are able to leverage the video for precise question answering, we compare our \emph{Vanilla SFT} model with two finetuned reference points in the social domain using \benchname. \emph{Question SFT} is trained to answer questions based solely on the question and answer text while \emph{Caption SFT} is its caption-based counterpart, which only has access to a question-independent, textual caption of the original video (generated by ASID-Captioner \cite{li2026universalvideomllmsattributestructured}) to answer corresponding questions. The results of \emph{Question SFT} show that despite the curation of \benchname, substantial learnable priors remain. Results using \emph{Caption SFT} lead to our third finding: \raisebox{-0.28\height}{\includegraphics[height=1.1\baselineskip]{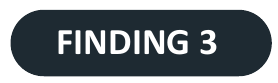}}~\textbf{A generic textual caption yields QA performance on par with processing the full multimodal data.} Although the experiment is limited to the social domain, the comparable performance in both settings illustrates the need for further research in understanding the ability of MLLMs to extract question-specific information from the audio and video stream. Finally, we discuss a practical application of our findings: since captions can be pre-computed independently of the questions, a two-stage approach utilizing captions for question answering enables very low latency at query time and multiple questions to be answered based on a single textual trace.

\section{Related Work}
\label{sec: related works}
\paragraph{\textbf{Multi-modal Large Language Models}}
After the rise of Large Language Models (LLMs) following the release of ChatGPT \cite{openai2022chatgpt} in 2022, researchers were quick to enable LLMs to process other modalities than text, starting with images. In \cite{liu2023visual} images are encoded and subsequently projected to the LLM token space, enabling strong LLM backbones to process images. This concept was quickly extended to the remaining modalities of audio and video \cite{zhang2023video}, at which point models able to process video, audio and text started to be called omni-models. Qwen2.5-Omni \cite{qwen2025qwen25omni} is an open-source omni-model that can output audio in addition to text. It is a common starting point for training specialized omni-models for various applications requiring access to all modalities \cite{yang2025humanomniv2, li2026universalvideomllmsattributestructured, luo2026unveiling}. 

\paragraph{\textbf{Reasoning models}}
At the start of 2025, DeepSeek-R1 \cite{guo2025deepseek} was released, demonstrating that the reasoning abilities of LLMs can be enhanced via a data-efficient reinforcement learning (RL) approach, making use of Group Relative Policy Optimization (GRPO) \cite{shao2024deepseekmathpushinglimitsmathematical}. 
%TO SAVE SPACE
%Previously, injecting LLMs with reasoning abilities was done via expensive supervised learning on human-produced examples. 
Previously, injecting LLMs with reasoning abilities was done via expensive supervised learning on human-produced examples. This new method caused an increase in the attention to reasoning and was quickly adapted to the MLLM domain \cite{zhou2025reinforced} by, among others, \cite{feng2026videor1, xiao2025perceptionr1, zhang2025rewatchr1}. The motivation for incorporating explicit thinking into these models is both explainability and to enhance downstream question-answering performance \cite{3600270.3602070}. However, explicit reasoning requires multiple forward passes through the network to generate additional reasoning tokens, resulting in a significant computational overhead. There are works claiming that the long reasoning chains that these thinking models produce are unnecessarily long, whereas better and quicker results can be obtained with shorter reasoning chains \cite{zhong2025rethinking, hassid2025dontoverthink}. AVATAR \cite{kulkarni2026avatar} addresses other limitations by incorporating an off-policy training architecture, improving sample efficiency, and a credit assignment strategy that emphasizes reasoning phases corresponding to planning and final answer synthesis. 

\paragraph{\textbf{Reasoning for social interaction understanding}}
The concept of explicit thinking or reasoning has also been extended to the social domain \cite{yang2025humanomniv2,qin2026humansense, li2025vegas, 11585891, luo2026unveiling, zhao2025r1omni, xiao2025perceptionr1}. A deep dive on reasoning in social settings is presented by the authors of \cite{mathur2025socialgenome}, providing a benchmark to evaluate the reasoning capabilities of MLLMs or omni-models. To the best knowledge of the authors, R1-Omni \cite{zhao2025r1omni}  is the first application of RL to an Omni-model for emotion recognition and the first application of RL  to such models for reasoning in the social setting in general. More recently, \cite{luo2026unveiling} introduced HitEmotion, a framework for emotion understanding based on Theory of Mind (ToM) modeling.\\ \\
HumanOmniV2 \cite{yang2025humanomniv2} focuses on the broader social setting and takes reasoning a step further. The authors identify two main issues, namely insufficient context understanding and shortcut reasoning. To address these issues, they enforce a sequential reasoning process in which the model first summarizes the relevant video context, then performs explicit reasoning based on that context, and finally generates its answer. They finetune Qwen2.5-Omni \cite{qwen2025qwen25omni} via a SFT training stage and implement two followup GRPO training phases. The resulting omni-model is able to answer questions about videos depicting social interaction but still generalizes to other settings. AffectOmni \cite{11585891} extends the HumanOmniV2 framework with \textit{People Focus} and \textit{Temporal Order} rewards, boosting performance on emotion recognition and temporally sensitive tasks. 

\section{\benchname}

%\steph{sounds incremental as a title}
\label{sec: IntentBench improvement}
\begin{figure}[tb]
  \centering
  \includegraphics[width=.9\textwidth]{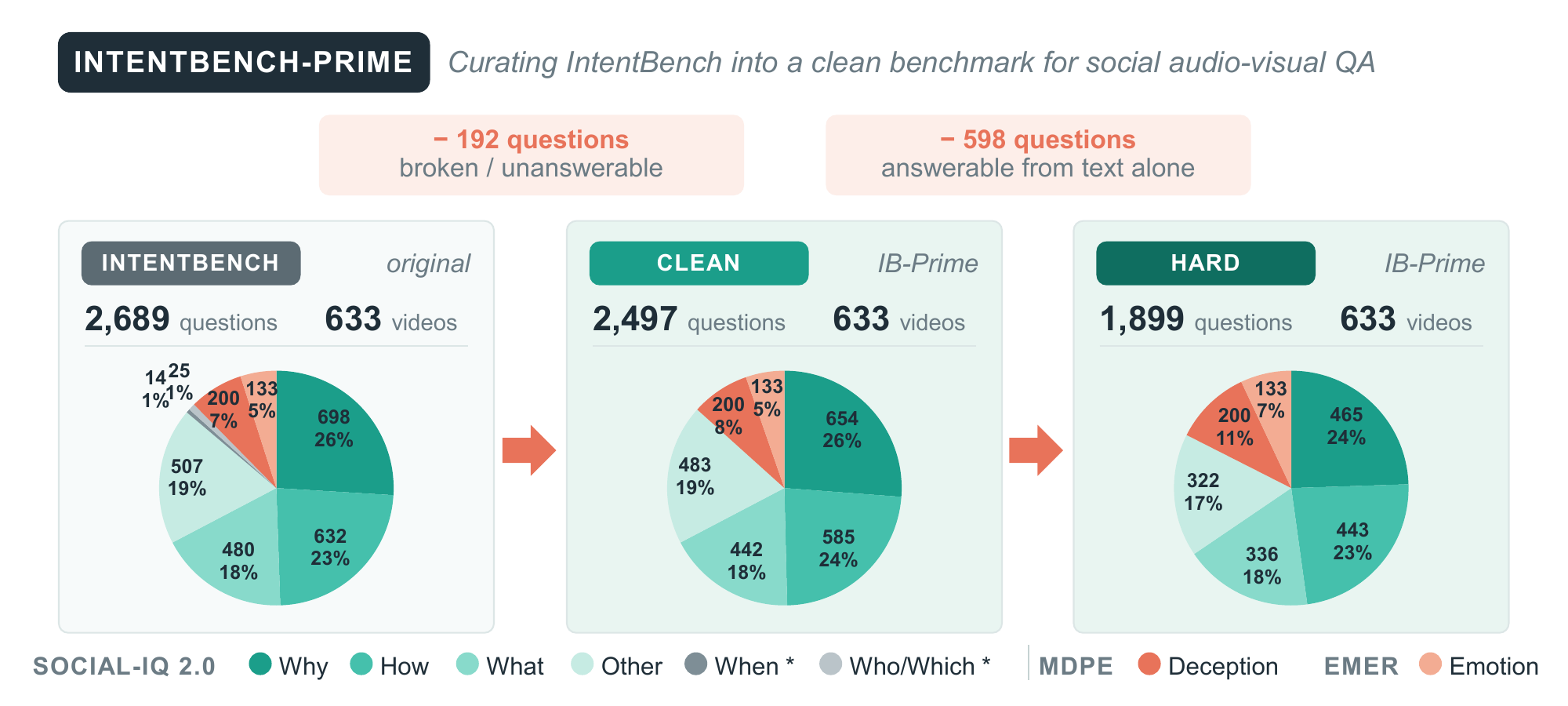}
  \vspace{-.1in}
  \caption{\textbf{Curation process of \benchname and benchmark statistics.} The two very small categories `when' (14) and `who/which' (25) are merged into `other' in both IntentBench-Prime variants. 616 questions are flagged as answerable from text alone; 18 of them were already removed as broken.}
  \label{fig: intentbench-prime statistics}
\end{figure}
\subsection{ \intentbench Diagnosis}
\begin{figure}[tb]
  \centering
  \includegraphics[width=\textwidth]{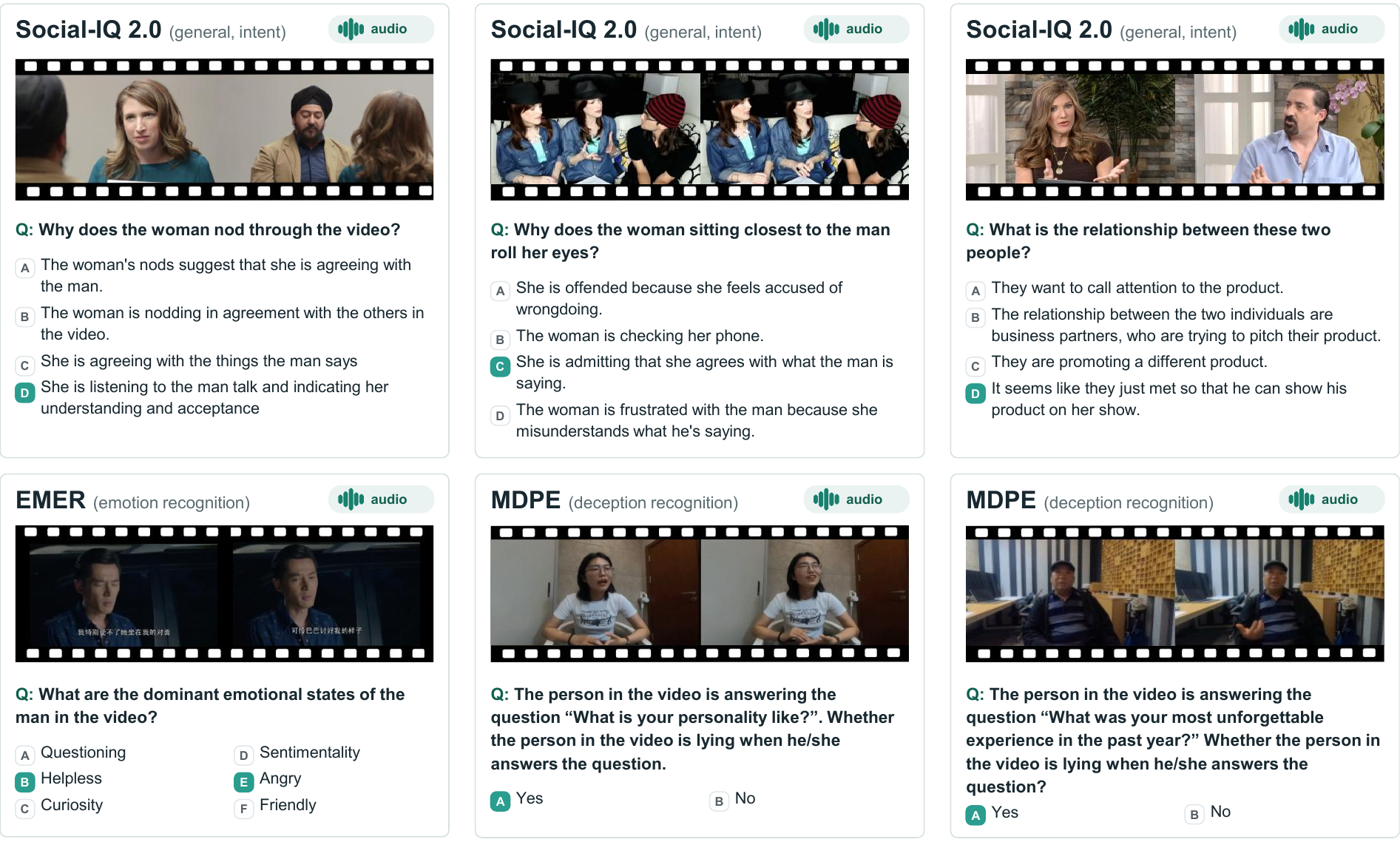}
  \vspace{-.3in}
  \caption{\textbf{Examples from Social-IQ 2.0, EMER and MDPE} as they are adapted in \intentbench. Images partly sourced from \cite{yang2025humanomniv2}.}
  \label{fig: examples data}
  
\end{figure}

%\paragraph{\textbf{Dataset Overview}}
Introduced by \cite{yang2025humanomniv2}, this benchmark evaluates the ability of omni-models to understand complex human social interactions from both video and audio. It consists of three categories derived from existing audio-visual datasets: intention and social intelligence (Social-IQ 2.0 \cite{zadeh2019social,siq2}), emotion recognition (EMER \cite{lian2023explainable}), and deception detection (MDPE \cite{cai2024mdpe}). Examples are shown in \cref{fig: examples data}. For the Social-IQ 2.0 subset, 300 videos with 2,356 manually verified multimodal questions are selected. GPT-4o with access to only the text modality is used to identify challenging questions and replace easy distractor answer options. From EMER, 133 videos are converted into six-option multiple-choice emotion recognition tasks, potentially with multiple correct answers, and evaluated using the F1-score. From MDPE, 200 videos containing truthful responses and varying levels of deception are selected, with the task requiring models to determine whether the speaker is lying. We identify three types of issues with the Social-IQ 2.0 (S-IQ) part of the benchmark, which, as displayed in \cref{fig: intentbench-prime statistics}, represents 87\% of the questions. Examples of issues are displayed in \cref{fig: issues IB} with more in the supplementary material, Appendix A. Below we provide details on each issue.\\

\begin{figure}[tb]
  \centering
  \includegraphics[width=.8\textwidth]{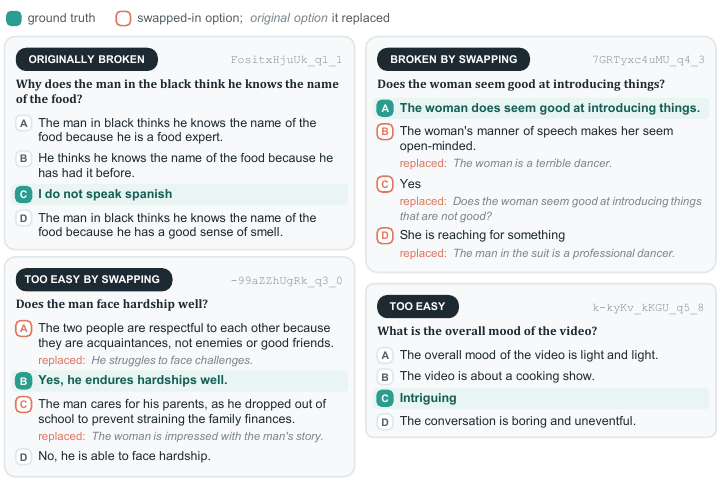}
  \vspace{-.1in}
  \caption{\textbf{Examples of issues in \intentbench}. More examples across all types of issues are provided in the supplementary material, Appendix A.}
  \label{fig: issues IB}
\end{figure}

\noindent\textit{\textbf{Issue 1: Text-only answerable questions}} \label{par: issue 1}
 We observe that a significant portion of questions can be answered with access to only the text modality (questions and answers), which means that they can be answered without the use of video and audio. To quantify this observation, we run four small-sized LLMs (Gemma-2-9B-Instruct \cite{gemma2technicalreport}, Llama-3.1-8B-Instruct \cite{llama3herdofmodels2024}, Mistral-7B-Instruct-v0.3 \cite{mistral7bv03release}, and Qwen3.5-9B \cite{qwen35technicalreport}) on only the S-IQ questions and answers of the benchmark. We propose to evaluate the text-answerability of the benchmark as the percentage of questions where all four LLMs agree on the correct answer, based on just the question and answers. This is a strict criterion, only capturing questions that we are sure are trivial. In practice, for a single model, text-only accuracy is higher as discussed in Section \ref{sec: IntentBench-Prime}. We run our analysis for \intentbench S-IQ questions, \intentbench S-IQ questions with original answer options (recall easy options are replaced for additional difficulty), and a set of S-IQ questions randomly selected from the original dataset, not in \intentbench, but the same size. We find that:
 %Note that for the emotion and deception categories, questions are always formulated the same way and hence nothing can be inferred from the question itself, which is why they are excluded from the analysis here. 
\begin{enumerate}
    \item  \intentbench Social-IQ 2.0 questions and \intentbench S-IQ questions with original answer options have similar text-answerability (26.1\% vs 24.4\%). This means that the replacement of distractor options to make questions harder in \intentbench was unsuccessful. We investigate this further and find that the replacement of easy options in  \intentbench is not done by creating new options, but by replacing distractor options with ground-truth answers from other questions corresponding to the same video. At least one distractor is swapped in 54\% of \intentbench Social-IQ 2.0 questions. The answer options from other questions can be semantically inconsistent with the question, seemingly making it easier for LLMs to identify them as not being the ground-truth answer.
    \item  \intentbench Social-IQ 2.0 questions with original answer options and a random set of S-IQ questions have similar text-answerability (24.4\% vs 24.2\%). This indicates that the curation process did not successfully select a harder subset of questions. After investigation of the question sets and corresponding videos, we find this to stem from a granularity mismatch in the selection process: filtering was done on the video level rather than on individual questions.
\end{enumerate}
%We investigate further and find three possible reasons for this finding. The main reason is that Social-IQ questions in  \intentbench are not sampled from the original data on a question-basis, but on a video basis\steph{what does it mean}. Combined with our results above it is clear that the usage of GPT-4o for the selection of hard questions was unsuccessful. 
%Second, the replacement of easy options in  \intentbench is not done by creating new options, but by replacing distractor options with ground-truth answers from other questions corresponding to the same video. At least one distractor is swapped in 54\% of  \intentbench Social-IQ 2.0 questions compared to the original. Note that this is not a feature of the original dataset\steph{not clear what you mean here}. The answer options from other questions can be semantically inconsistent with the question, making it easier to for LLMs to identify them as not being the ground-truth answer.
Additionally, in HumanOmniV2, GPT-4o with only text modality is used to identify difficult questions and presumably also to replace easy options. To measure the increase in difficulty due to the modifications, HumanOmniV2 also uses GPT-4o. Using the same model for both of these tasks likely gave an inflated measure of increased difficulty. \\

\noindent\textit{\textbf{Issue 2: Broken questions.}}
We further investigate the swapping of distractor options with ground-truth answers from other questions. In addition to the fact that, contrary to the intentions of the constructors, the benchmark becomes slightly easier for LLMs, we find that the option-swapping can also break questions. For example, \cref{fig: issues IB} shows an example of a question where the swapped-in option is equal to the ground-truth of the original question. Another relatively common problem is that the ground-truth can be a sub- or super-string of a distractor option, in the form of e.g. \textit{A. Yes} and \textit{B. Yes, [reason / rephrasing of question]}, which also makes the correct ground-truth ambiguous. \\ \\
An important note is that, for omni-models, the Social-IQ questions with swapped options are harder. This is shown in \cref{tab: IB swapping} where the 1,264 questions for which options were swapped are isolated in the two right-most columns. It is clear that despite the change having little effect for the LLMs w.r.t. overall difficulty, there is an effect for Qwen2.5-Omni and HumanOmniV2 models. The reason for this is likely that the swapped-in options are GT answers to another question, meaning they are factually grounded in the video and audio.\\
%Note also that the set of questions for which options were changed were, not unexpectedly, originally some of the easier questions as seen by the significantly above-average performance on this part of the benchmark, even after the change\steph{I don't get this sentence}. 

\begin{table}[tb]
  \caption{Average accuracy on all IB Social-IQ questions, all IB Social-IQ questions where an option is swapped, and the same questions but with the swapping undone.}
  \label{tab: IB swapping}
  \scriptsize
  \centering
  \begin{tabular}{@{}ll|ccc@{}}
    \toprule
    \multicolumn{2}{@{}l|}{\textbf{Benchmark subset} $\rightarrow$} &  \quad IB S-IQ full \quad & \quad IB S-IQ swapped \quad & \quad IB S-IQ unswapped \quad \\
    \multicolumn{2}{@{}l|}{\textbf{Number of samples} $\rightarrow$}               & 2,356 & 1,264 & 1,264 \\
    \midrule
    \textbf{Method $\downarrow$} & \textbf{Input} & & & \\ 
    \midrule
    Qwen2.5-Omni-7B \cite{qwen2025qwen25omni} & A+V+T & 64.5 & 74.8 & 84.7\\
    HumanOmniV2 \cite{yang2025humanomniv2}     & A+V+T & 68.9 & 75.9 & 88.1 \\
    LLMs agree on GT                         & T & 26.1 & 42.4 & 39.2 \\
  \bottomrule
  \end{tabular}
\end{table}

\noindent\textit{\textbf{Issue 3: Original dataset issues.}}
We also discover other issues in \intentbench, that after investigation are determined to stem from the original dataset, such as a ground truth that makes no sense ("this is a test", "I do not speak Spanish", "N/A (Wrong video clip)"), questions or ground-truth answers with a major typo, and questions with two (almost) identical answer options of which one is GT. These broken questions amount to almost 6\% of Social-IQ questions. Although not investigated in more detail in this paper, this number is a proxy for the amount of broken questions in the Social-IQ 2.0 dataset in general.
%The two previously discussed problems add up into a significant amount of broken questions across the benchmark, which is obviously highly undesirable. It causes comparisons between models with small performance differences to be invalid.

\subsection{\benchname }
\label{sec: IntentBench-Prime}
%Some of our findings contradict the goals the original authors had in mind for  \intentbench and other findings are clearly undesirable in a benchmark. The identified problems lead to a significant amount of noise in  \intentbench, making finegrained comparison of models unreliable. Hence, we address these problems by:
Motivated by the discovery of these issues, we propose a curated version of \intentbench that we name \benchname. It is curated using the following steps:
\begin{enumerate}
    \item \textbf{Removing broken questions from the benchmark}, meaning questions with any of the breaking problems that were discussed prior. This is done by making a ranking of the questions, placing questions with highest suspicion of being broken at the top. The ranking consists of three elements, for which more details can be found in the supplementary material, Appendix B.
    %TO SAVE SPACE
    %is based on programmatic checks (to identify for example two identical answers of which one is GT), an audit of Claude Haiku 4.5 \cite{anthropic2026claudehaiku}, and performance of omni models with and without access to an oracle hint for each question. Details can be found in the supplementary material, Appendix B. 
    We manually go through all Social-IQ 2.0 QA pairs in the benchmark in order of the ranking. We only inspect the video when we highly suspect a question is broken but cannot conclusively determine it from the QA pair alone. When we do not identify a broken question for 300 questions in a row, we stop the cleaning procedure. At this point, more than half of the S-IQ questions have been manually verified. With our pipeline, 192 out of 2,356 S-IQ questions in \intentbench are removed. Of these 192, 57 (30\%) were questions with swapped distractor options, the rest being issues with the original dataset. 
    \item \textbf{Removing questions that are answerable without looking at the video / audio.} Here, we drop the questions that are identified by the ensemble of 4 LLMs in Section \ref{par: issue 1} to be text-answerable. That means, when all 4 LLMs are able to identify the correct answer to a question based on just the questions and answers, we remove this question from the benchmark. We utilize this strict criterion because we aim to be conservative and retain non-trivial questions. Using this method, we are able to exclude 616 trivial questions from the benchmark. Out of these 616, 536 (87\%) were questions with swapped distractor options. There is an overlap of 18 questions between the broken and text-answerable sets. 
\end{enumerate}
We do not undo the swapping of distractor options entirely despite it being an unusual and as shown, sometimes breaking, change. The reason for this is twofold: first of all, the omni-models seemingly struggling with this change makes it an interesting property of the benchmark. Secondly, many of the questions with options swapped are present in the two sets of removed questions listed above, meaning the questions that remain should be fair questions that are a challenge for omni-models, which is obviously a desired property for this benchmark. Note that we do not identify problems with the emotion and deception categories of the benchmark and leave those intact. Finally, two very small Social-IQ 2.0 sub-categories in the original  \intentbench, `when' (14 samples) and `who/which' (25 samples) are merged into the `other' sub-category. \\ \\
%TO SAVE SPACE
%Finally, there are two very small Social-IQ 2.0 sub-categories in the original  \intentbench, `when' (14 samples) and `who/which' (25 samples), which we merge into the `other' sub-category as we deem performance measured on such small categories to be statistically insignificant.
\noindent Since \benchname is constructed strictly by removal of questions as opposed to modification or addition of questions, its release consists of an exclusion list. Any existing method with per-question results on \intentbench can obtain \benchname scores by simply filtering its existing results, meaning no re-evaluation runs are required, a highly practical property. 
\begin{table}[t]
  \caption{Performance on the different versions of \benchname. Detailed performance across categories can be found in the supplementary material, Appendix C. Granite-4.1-8B is an LLM answering the questions based on just the QA pairs.}
  \label{tab: new IB}
  \scriptsize
  \centering
  \begin{tabular}{@{}ll|ccc@{}}
    \toprule
    \multicolumn{2}{@{}l|}{\multirow{2}{*}{\textbf{Statistics} $\downarrow$ \quad \textbf{Benchmark} $\rightarrow$}} & \multirow{2}{*}{\quad \intentbench \quad} & \multicolumn{2}{c}{\benchname} \\
    \cmidrule{4-5}
    \multicolumn{2}{@{}l|}{} & &  Clean & Hard \\
    \midrule
    \multicolumn{2}{@{}l|}{\# questions} & 2,689 & 2,497 & 1,899 \\
    \multicolumn{2}{@{}l|}{\# videos}    & 633   & 633   & 633   \\
    \midrule
    \textbf{Method} $\downarrow$ & \textbf{Input} &       &       &       \\
    \midrule
    Qwen2.5-Omni-7B \cite{qwen2025qwen25omni}  & A+V+T & 64.6 (64.2*) & 67.8 & 63.2 \\
    HumanOmniV2 \cite{yang2025humanomniv2}     & A+V+T & 68.9 (69.3*) & 71.8 & 66.9 \\
    Qwen2.5-Omni-7B                            & T (QA only) & 55.1 & 57.9 & 46.9 \\
    Granite-4.1-8B \cite{ibm2026granite41}     & T (QA only) & 52.1 & 53.7 & 41.5 \\
  \bottomrule
  \addlinespace[2pt]
  \multicolumn{5}{@{}l@{}}{\scriptsize $^{*}$ reported from \cite{yang2025humanomniv2}. Other numbers are computed using our pipeline.} 
  \end{tabular}
  \vspace{-0.2in}
\end{table}

\noindent\textbf{\textit{Quantitative analysis of \benchname.}}
Question count, video count, and performance of Qwen2.5-Omni and HumanOmniV2 for three versions of  \intentbench are shown in \cref{tab: new IB}: the original,  \intentbench with broken questions removed (\benchname (Clean)), and finally  \intentbench with broken and trivial questions removed (\benchname (hard)). Note that the emotion category is scored with F1-score, and that all categories are micro-averaged to obtain the overall accuracy. We also report the performance of Qwen2.5-Omni with only text as input to quantify the drop in text-answerability. Its accuracy of 46.9\% on the hard split remains above the random-chance level ($\sim$28\%), indicating that text-answerability is not fully eliminated. To complete the analysis, we include a text-only LLM (Granite-4.1-8B~\cite{ibm2026granite41}) to assess if a more recent and text-specialized model could improve performance. We surprisingly observe the opposite trend with Granite underperforming Qwen2.5-Omni text-only baseline by 5.4\% on the \benchname (Hard) set. We hypothesize this result might be explained by either the benefit of multimodal training or simply differences in training dataset distribution.  Nevertheless, we retain our strict removal criterion (unanimous agreement of all four LLMs), as loosening it, \eg to a 3-out-of-4 majority, would also discard questions that remain challenging for omni-models and would further shrink the benchmark. Additional statistics for all three benchmark versions are  displayed in \cref{fig: intentbench-prime statistics}.

\section{Ineffectiveness of Reasoning for Social AV-QA}
\label{sec: reasoning versus direct AVQA}
\subsection{Experiment setup}
\label{subsec: experimental setup}
The goal is to identify whether training a model to generate a reasoning trace improves performance on audio-visual question answering compared to training a model to directly answer the questions, with no intermediate steps. To isolate this result, we finetune the same base model (Qwen2.5-Omni) and train on the exact same original audio-video / image QA data used in HumanOmniV2. The model is trained to directly answer questions based on the audio and video with no intermediate steps, outputting only the letter(s) corresponding to the perceived correct answer(s).\\
For sufficient benchmark diversity, we evaluate on \worldsense \cite{hong2026worldsense}, \dailyomni \cite{zhou2025daily}, \intentbench, and \benchname. The latter two are considered in-domain whereas \worldsense and \dailyomni are out-of-domain with respect to the training data. 
\noindent\textbf{\worldsense} \cite{hong2026worldsense} is a benchmark with 3,172 questions corresponding to a diverse collection of 1,662 videos which include audio. The benchmark is designed to demand a strong coupling between video and audio, requiring strong cross-modal alignment to perform well. \textbf{\dailyomni} \cite{zhou2025daily} is a similar benchmark, also focusing on both video and audio and the corresponding cross-modal connections. It has 684 real-world videos with 1,197 corresponding questions. There are six categories of which two are particularly interesting in the context of this paper: context understanding and reasoning. Additionally, there are 30 and 60-second duration based subsets. \\ \\
%TO SAVE SPACE
%\noindent\textbf{\worldsense} \cite{hong2026worldsense} is a benchmark with 3,172 questions corresponding to a diverse collection of 1,662 videos which include audio. The benchmark is designed to demand a strong coupling between video and audio, requiring strong cross-modal alignment to perform well. It is of high quality, being annotated by 80 experts over multiple rounds of correction. \noindent\textbf{\dailyomni} \cite{zhou2025daily} is a similar benchmark, also focusing on both video and audio and the corresponding cross-modal connections. It has 684 real-world videos with 1,197 corresponding questions. There are six categories of which two are particularly interesting in the context of this paper: context understanding and reasoning. Additionally, there are 30 and 60-second duration based subsets. \\ \\
To validate our evaluation pipeline is working as intended we first reproduce the benchmarks' Qwen2.5-Omni baseline numbers and the numbers from the HumanOmniV2 paper. Differences in our reproduction numbers are small, and are suspected to stem from slight prompting differences and FPS and maximum allowed frames settings. \\ \\
\noindent\textit{\textbf{Vanilla SFT training setup.}} For the training data, we aim to mirror the exact training data used in HumanOmniV2 \cite{yang2025humanomniv2} for a fair comparison. This includes data from OmniInstruct \cite{li2025omnibenchfutureuniversalomnilanguage}, Video-R1 \cite{feng2026videor1}, Social-IQ 2.0 training data, and the 200 entries from EMER not present in \intentbench. We follow the HumanOmniV2 code-base to match the training data, which results in 20K videos and 10K images and their corresponding multiple-choice questions and answers being used.
%find inconsistencies between the paper-reported training data and what is used based on their codebase \steph{can we add examples in parentheses}\koen{maybe just remove instead. It is quite insignificant for training, nobody cares, and nobody will ever find out. Just say we follow as closesly as possible from codebase}, so we follow the codebase as closely as possible, assuming the setup indicated there was used to produce the results. The training data we use contains 20K videos and 10K images and their corresponding multiple-choice questions and answers. \\ \\  
We finetune the thinker part of Qwen2.5-Omni-7B \cite{qwen2025qwen25omni}, freezing the video \& audio encoders and aligner layers. We train LoRA with learning rate $1e^{-4}$ for 1 epoch. LoRA rank is 16 and the video settings are 2 FPS with 32 maximum frames, evenly spreading when the video is longer than 16 seconds. This mirrors the setup of evaluation in HumanOmniV2. Training takes less than 4.5 hours on 4$\times$ H100 80 GB. We also train a model using full parameter finetuning, which takes 2 epochs to reach similar or slightly worse performance than LoRA on the benchmarks. 

\subsection{Results and Discussions}
\begin{table*}[t]
%\vspace{-.2in}
\centering
\caption{Performance comparison on all evaluated benchmarks.}
  \vspace{-.2in}
\label{tab:main_results}
\begin{subtable}{\textwidth}
\centering
\caption{\intentbench and \benchname. The `Intent' category corresponds to all Social-IQ 2.0 questions. Emotion and Deception are unchanged across variants (no such questions are removed) and are therefore reported only once, under \intentbench. Detailed results can be found in the supplementary material, Appendix C.}
\scriptsize
\begin{tabular}{@{}lc|cccccccc@{}}
\toprule
\multicolumn{2}{@{}l|}{\multirow{2}{*}{\textbf{Benchmark} $\rightarrow$}}
  & \multicolumn{4}{c}{} & \multicolumn{4}{c}{\benchname} \\
\cmidrule(lr){7-10}
\multicolumn{2}{@{}l|}{}
  & \multicolumn{4}{c}{\intentbench}
  & \multicolumn{2}{c}{\textbf{Clean}}
  & \multicolumn{2}{c}{\textbf{Hard}} \\
\cmidrule(lr){3-6} \cmidrule(lr){7-8} \cmidrule(lr){9-10}
\textbf{Method} & \textbf{CoT} \quad 
  &  Intent  \quad  &  Emo.$^{*}$  \quad  &  Dec.$^{**}$  \quad &  \textbf{Avg.} \quad  
  &  \quad Intent \quad  &  \textbf{Avg.} \quad 
  &   Intent \quad  & \textbf{Avg.} \quad  \\
\midrule
Qwen2.5-Omni-7B \cite{qwen2025qwen25omni}    & $\times$   & 64.5 & 71.4 & 61.2 & 64.6 & 68.2 & 67.8 & 62.8 & 63.2 \\
AVATAR (CVPR 2026) \cite{kulkarni2026avatar} & \checkmark & -    & -    & -    & 63.9 & -    & -    & -    & -    \\
AffectOmni (IEEE TAC 2026) \cite{11585891}       & \checkmark & 69.7 & \textbf{85.4} & 62.5 & 69.9 & - & - & - & - \\
HumanOmniV2 \cite{yang2025humanomniv2}       & \checkmark & 69.1 & 82.4 & \textbf{64.0} & 69.3 & 72.2 & 71.8 & 66.5 & 66.9 \\
Vanilla SFT (Full FT)                        & $\times$   & 70.8 & 82.6 & 58.2 & 70.5 & 74.4 & 73.6 & 69.2 & 69.0 \\
Vanilla SFT (LoRA)                           & $\times$   & \textbf{70.9} & 83.2 & 58.2 & \textbf{70.6} & \textbf{74.6} & \textbf{73.8} & \textbf{70.9} & \textbf{70.4} \\
\bottomrule
\addlinespace[2pt]
\multicolumn{10}{@{}l@{}}{\scriptsize $^{*}$Emo. = Emotion \quad $^{**}$Dec. = Deception} \\
\end{tabular}

\end{subtable}

\vspace{0.8em}

\begin{subtable}{\textwidth}
\centering
\caption{\worldsense and \dailyomni. We include some of the interesting subcategories of \dailyomni.}
\scriptsize
\begin{tabular}{@{}lc|cccccc@{}}
\toprule
\textbf{Benchmark $\rightarrow$}
& & \quad \multirow{2}{*}{\worldsense} \quad \quad
& \multicolumn{5}{c}{\dailyomni} \\
\cmidrule(lr){4-8}
\textbf{Method} & \textbf{CoT} \quad
& & CU$^{*}$ & Reasoning & \quad 30s \quad  & \quad 60s \quad & \quad \textbf{Avg.} \quad \\
\midrule
Qwen2.5-Omni-7B  \cite{qwen2025qwen25omni}  & $\times$  & 45.4 & 58.6 & 73.1 & 64.6 & 59.1 & 62.1 \\
AVATAR (CVPR 2026) \cite{kulkarni2026avatar} &  \checkmark & 46.0 & - & - & - & - & 55.7 \\
AffectOmni (IEEE TAC 2026) \cite{11585891}                  & \checkmark & \textbf{48.8} & 56.0 & \textbf{78.2} & - & 58.6 & 61.9 \\
HumanOmniV2  \cite{yang2025humanomniv2}     & \checkmark  & 47.1 & 51.8 & 74.3 & 63.1 & 53.1 & 58.5 \\
Vanilla SFT (Full FT)                       & $\times$  & 46.7 & 57.0 & 76.0 & 65.2 & 58.4 & 62.1 \\
Vanilla SFT (LoRA)                          & $\times$   & \textbf{48.8} & \textbf{60.1} & 76.6 & \textbf{68.3} & \textbf{61.4} & \textbf{65.2} \\
\bottomrule
\addlinespace[2pt]
\multicolumn{8}{@{}l@{}}{\scriptsize $^{*}$CU = Context Understanding} 
\end{tabular}
\end{subtable}
\end{table*}

Results for \intentbench, \benchname, \worldsense, and \dailyomni are shown in \cref{tab:main_results}. For HumanOmniV2 \cite{yang2025humanomniv2} on \intentbench, \worldsense, and \dailyomni, we report not our reproduction numbers but the original paper numbers, as the latter were generally marginally higher. This way, our comparison is conservative with respect to our claims. We include results on \intentbench despite its known limitations, as it enables comparison with methods whose results we could not reproduce. In the deception category, 4 out of 200 videos led to a loading error, hence those are excluded from the evaluation.\\ \\
\noindent As can be seen in \cref{tab:main_results} (a) and (b), \emph{Vanilla SFT}, despite its simple training setup, consistently outperforms both the baseline and HumanOmniV2 numbers. The one exception is the deception category, where HumanOmniV2 reports 64.0\%. However, our reproduction of HumanOmniV2 numbers reaches only 60.2\% on this category. We note all methods obtain close to chance on this small binary subset. The most striking result is on \dailyomni, where HumanOmniV2 performance degrades compared to the baseline. Interestingly enough, our baseline also outperforms on the Context Understanding and Reasoning categories of \dailyomni, despite these categories seemingly being particularly well-suited to the HumanOmniV2 model. Additional models leveraging reasoning, AVATAR \cite{kulkarni2026avatar} and AffectOmni\footnote{AffectOmni reports a macro-average over all categories of \intentbench, which leads to a different number. Due to the imbalance between categories and to remain consistent with other methods, we deem micro-average the most suitable metric.} \cite{11585891}, are not able to outperform our simple baseline, trained with HumanOmniV2 data, on any of the benchmarks. AffectOmni, which is trained with the same HumanOmniV2 data, does take the top spot in some of the sub-categories but remains worse on average. To our knowledge, there is no reasoning method with a comparable base-model and size that outperforms our \emph{Vanilla SFT} baseline on the discussed benchmarks. \\ \\
The impressive performance of \emph{Vanilla SFT} suggests that, whenever possible, this simple baseline should be included as standard procedure in future papers on audio-visual (social) question answering. This way, benchmark gains stemming from fine-tuning on the task using training data can be properly disentangled from gains originating from a superior method. To our knowledge, direct supervised finetuning with LoRA as in our \emph{Vanilla SFT} baseline is the simplest and cheapest way one can finetune a model, and hence obtain such a baseline. \\ \\ %Compared to training a model to reason, it is cheaper data-wise (no need to distill traces from a proprietary model), significantly faster and inexpensive to train, and drastically quicker at inference time.\\
\noindent \textbf{\textit{Computational cost and latency analysis.}}
In addition to its accuracy, \emph{Vanilla SFT} benefits from several practical advantages in compute cost and inference latency compared to reasoning models. We now provide a quantitative analysis of these gains. To estimate the training cost of HumanOmniV2 training, we use training time and trace statistics measured from internal training runs with SFT and GRPO on long traces, making a conservative estimate of the total time. Results are reported in \cref{tab: latency}. Overall, it clearly demonstrates the superiority of the reasoning-free approach in terms of compute cost both at training and test time. The gains are especially clear in terms of decoding latency (356$\times$ factor gain) as direct answering avoids the auto-regressive cost of CoT. 

% as it shows that reasoning is ineffective for improving performance for audio-visual QA. Therefore, we strongly advocate for usage of our proposed baseline in papers in this domain. Finally, one possible advantage of reasoning traces is increased interpretability of the model's answers, but whether or not this is true or significant is outside the scope of this paper. 
\begin{table}[tb]
  \caption{Inference efficiency on the full \benchname-Hard split.
  Same hardware (1$\times$H100), where necessary, native protocol for the models is used, other than that the setup is identical. 
  The latency gap is fully attributed to the decode stage, i.e. the auto-regressive generation of the output trace.
  }
  \label{tab: latency}
  \scriptsize
  \centering
  \begin{tabular}{@{}l|cc|ccc|c@{}}
    \toprule
    & \multicolumn{2}{c|}{\textbf{Cost}} & \multicolumn{3}{c|}{\textbf{Latency/question (s)}} & \textbf{Accuracy (\%)}\\
    \cmidrule(lr){2-3}\cmidrule(lr){4-6}
    \textbf{Method} & Train & Output & Prefill & Decode & Total & IB-Prime (Hard)\\
                    & (GPU\,h) & (tok.) & & & & \\
    \midrule
    Direct answering (Vanilla SFT)                        & \textbf{18} & \textbf{2} & 1.22 & \textbf{0.02} & \textbf{1.23} & \textbf{70.4}\\
    Reasoning (HumanOmniV2 \cite{yang2025humanomniv2}) &  $\sim$200       & 527          & \textbf{1.19} & 7.12          & 8.31          & 66.9\\
    \midrule
    Reasoning \emph{vs.} direct answering                  &    $\sim$11$\times$      & $263\times$  & $1.0\times$ & $356\times$ & $6.7\times$ & $-3.5$\\
    \bottomrule
  \end{tabular}
\end{table}
%A recent work also based on HumanOmniV2, MODF-SIR, seems to surpass our baseline for the WorldSense benchmark, but we exclude this model from our analysis as we find severe problems with the accuracy calculations in their work\footnote{Shortly, it seems pass@k accuracy is reported as plain accuracy without mention.}. \koen{I don't know if it is okay to call out a paper like this (it is also not at a conference so who cares) and I guess this is a lot less important than just adding more other methods in the tables, but I feel the need to mention this in case a reviewer looks up other work based on HumanOmniV2}. 

\section{Limited Gap Between Video and Caption-Based Answering}
\label{sec: when to use reasoning traces}
Motivated by the results of Section \ref{sec: reasoning versus direct AVQA}, we aim to understand more deeply to what extent MLLMs are able to leverage the video modality to provide accurate answers. To investigate this, we introduce two reference points. First, we introduce \emph{Question SFT}, fine-tuning Qwen2.5-Omni to answer the questions exclusively from the QA text, without any context. This serves as a context-free reference, depicting the performance a finetuned MLLM can achieve by learning question-answer structure and other dataset priors. Secondly, we introduce \emph{Caption SFT}, fine-tuning Qwen2.5-Omni to answer questions using only a textual caption of the corresponding video. Captions are generated using the recent ASID-Captioner model~\cite{li2026universalvideomllmsattributestructured}. The captions are generated independently of the questions and are therefore not conditioned on the information required to answer them. These baselines are compared with the \emph{Vanilla SFT} method from Section~\ref{sec: reasoning versus direct AVQA}. We limit the training data to Social-IQ 2.0 and EMER to reduce the cost of the experiment. Apart from the modality setup, training is identical between the three variants. Evaluation is done using \benchname (Hard) in the same modality setting as training, for each model respectively. \\
\begin{table}[tb]
  \caption{Comparison of Qwen2.5-Omni-7B (top) in various modality settings to models finetuned (bottom) in those settings on \benchname (Hard). The three settings are: answering questions based on 1. Audio and video (\emph{Vanilla SFT}), 2. A finegrained but question-independent caption generated by ASID-Captioner \cite{li2026universalvideomllmsattributestructured} (\emph{Caption SFT}), 3. Only the question and answer pair (\emph{Question SFT}).
  }
  \label{tab: vanilla sft vs caption sft}
  \scriptsize
  \centering
  \begin{tabular}{@{}ll|ccccccc@{}}
    \toprule
    \multicolumn{2}{@{}l|}{\textbf{Method} $\downarrow$} & \multicolumn{7}{c}{\textbf{Benchmark:} \benchname (hard)} \\    
    \midrule
    \textbf{Model} & \textbf{Modalities} \quad & Why & How & What & Other & Emo.* & Dec.**  & \textbf{Avg.} \\
    \midrule
    Qwen2.5-Omni-7B & T (Q+A only) & 44.1 & 43.3 & 43.5 & 55.3 & 37.7 & 60.0 & 46.9 \\
    Qwen2.5-Omni-7B & T (ASID caption \cite{li2026universalvideomllmsattributestructured}) & 62.6 & 56.9 & 61.6 & 65.2 & 70.0 & 63.5 & 62.1 \\
    Qwen2.5-Omni-7B \quad \quad & A+V+T & 61.7 & 60.3 & 65.8 & 64.6 & 71.4 & 61.2 & 63.2 \\
    \midrule
    Question SFT & T (Q+A only) & 61.9 & 58.2 & 57.1 & 68.0 & 75.8 & 60.0 & 62.0\\
    Caption SFT & T (ASID caption \cite{li2026universalvideomllmsattributestructured}) & 68.2 & 64.1 & 68.5 & 74.2 & 84.5 & 60.5 & 68.6 \\
    Vanilla SFT & A+V+T & 68.6 & 67.0 & 67.6 & 74.5 & 86.8 & 60.7 & 69.5 \\
    \bottomrule
  \addlinespace[2pt]
\multicolumn{9}{@{}l}{\scriptsize $^{*}$Emo. = Emotion \quad $^{**}$Dec. = Deception}  
\end{tabular}
\end{table}

\noindent Results in \cref{tab: vanilla sft vs caption sft} show Qwen2.5-Omni baseline results for the three settings (top) and their finetuned counterparts (bottom). Results show that the finetuned \emph{Question SFT} model achieves performance on the benchmark 15.1\% higher than its Qwen2.5-Omni baseline. Part of this gain is to be expected, as training on same-source QA data increases format alignment and task knowledge, even without context. However, the magnitude of the gain is more than twice that of the context-based variants. Furthermore, the resulting model comes within 7.5\% of \emph{Vanilla SFT} and 6.6\% of \emph{Caption SFT}, indicating that there is substantial structure being learned in Social-IQ 2.0 and EMER questions. We note that this prior learning cannot explain the out-of-domain gains of \emph{Vanilla SFT} on WorldSense and Daily-Omni reported in Section \ref{sec: reasoning versus direct AVQA}. \\ \\
Turning to the contribution of context, in both the un-finetuned (top in \cref{tab: vanilla sft vs caption sft}) and finetuned setting (bottom), caption-based answering performance is on par with video-based answering, the difference being close to a single percentage point in both settings. This finding implies that the video-based \emph{Vanilla SFT} model, despite its state-of-the-art performance, is unable to extract substantially more information from the video than its caption-based counterpart from a textual caption. Although the ASID-Captioner model \cite{li2026universalvideomllmsattributestructured}, which shares Qwen2.5-Omni as base model, has been trained to generate a fine-grained textual caption based on the video and audio, it is a general-purpose caption model not geared towards the social domain. Importantly, the captions are generated unconditioned on the question. These results motivate further research to be done into the ability of MLLMs to extract question-relevant information from videos. \\  
\noindent\textbf{Potential applications of textual traces.} The remarkably high performance of the introduced caption-only baseline motivates the usage of textual captions as a replacement for the corresponding video. A potential setting where the latter can be useful is a two-stage approach where an MLLM is first used to generate a dense caption of the video, independent of possible following questions, and then answering based on the generated textual trace. Although the generation of the caption is a computational burden, this enables the video to be `pre-processed', decoupling the heavy video token processing from question answering. This drastically reduces inference time from the moment the query arrives. Additionally, as the caption is question-independent, it can be used to answer multiple questions.

\section{Conclusion}
In this paper, we present three key findings regarding the social understanding capabilities of Multimodal Large Language Models (MLLMs). First, we identify several issues within \intentbench and address them through a combination of automated and manual filtering. This pipeline yields \benchname, an improved benchmark for analyzing the social understanding of omni-models. Second, we investigate the effectiveness of reasoning for (social) audio-visual QA. By applying standard supervised fine-tuning, we introduce a baseline model, \emph{Vanilla SFT}, that directly answers questions. Despite its simplicity, it achieves state-of-the-art performance among comparable models, outperforming or matching complex CoT reasoning methods. Consequently, we advocate for establishing standard SFT as a mandatory baseline to accurately measure methodological progress in the field. Finally, we examine the extent to which \emph{Vanilla SFT} actually relies on audio-visual input. Our analysis reveals that models can learn substantial priors solely from the text modality in the training datasets. Additionally, a fine-grained but question-independent textual caption yields performance on par with Vanilla SFT. Together, these findings underscore the need for further research into how effectively MLLMs extract question-specific information directly from video modalities.%Simultaneously, our results motivate the usage of textual traces in a two-stage framework.      

\paragraph{\textbf{Limitations.}} 
For practical reasons, our curation of \benchname from \intentbench is limited to the removal of questions as opposed to addition or modification of questions. This reduces the amount of samples in the benchmark and only lightly balances the category distribution of the benchmark. Furthermore, while \benchname (Hard) excludes a significant chunk of questions answerable by pretrained LLMs, our \emph{Question SFT} reference shows that a model finetuned on data from the same source as the benchmark can still exploit considerable priors. %For our investigation into the effectiveness of reasoning and the proposal of our \emph{Vanilla SFT} baseline, results remain limited to the one base model (Qwen2.5-Omni) and a single training dataset. However, adaptation of our proposed baseline in future papers would naturally expand the scope of these results. 
Additionally, across all experiments, our results are limited to a single base model, Qwen2.5-Omni.
Finally, the comparison of video-, caption- and question-based answering remains limited to the social domain and further investigation can be done to fully disentangle these results into format alignment, source-specific priors, or potential other sources. We deem the result an interesting starting point for multiple avenues of future research.

\section*{Acknowledgements}
This work was granted access to the HPC resources of IDRIS under the allocation 2026-AD011017620 made by GENCI. \\ \\ 
E. Talavera Martinez was supported by the NWO Talent Programme – VENI (project Understanding Social Interactions in First-Person Videos with Multimodal Learning, file number 244507) which is financed by the Dutch Research Council (NWO).

%\clearpage
\bibliographystyle{splncs04}
\bibliography{main}

@String(CVPR  = {IEEE Conf. Comput. Vis. Pattern Recog.})

@String(ICLR  = {Int. Conf. Learn. Represent.})

@String(AAAI  = {AAAI})

@String(TMLR  = {Trans. Mach. Learn Res.})

@String(CVPR  = {CVPR})

@String(ICLR  = {ICLR})

@String(TMLR  = {TMLR})

@misc{ibm2026granite41,
  author       = {{IBM Research}},
  title        = {Granite 4.1 Language Models},
  year         = {2026},
  howpublished = {\url{https://www.ibm.com/granite/docs/models/granite4-1}},
}

@article{llama3herdofmodels2024,
  title   = {The Llama 3 Herd of Models},
  author  = {{Llama Team, Meta}},
  journal = {arXiv preprint arXiv:2407.21783},
  year    = {2024},
  url     = {https://ai.meta.com/blog/meta-llama-3-1/}
}

@article{gemma2technicalreport,
  title   = {Gemma 2: Improving Open Language Models at a Practical Size},
  author  = {{Gemma Team} and Riviere, Morgane and Pathak, Shreya and Andreev, Alek and others},
  journal = {arXiv preprint arXiv:2408.00118},
  year    = {2024}
}

@article{qwen35technicalreport,
  title   = {Qwen3.5 Technical Report: Towards Native Multimodal Agents},
  author  = {{Qwen Team}},
  journal = {arXiv preprint arXiv:2604.15804},
  year    = {2026}
}

@misc{mistral7bv03release,
  author       = {{Mistral AI Team}},
  title        = {Mistral-7B-Instruct-v0.3 Open-Weight Model},
  year         = {2024},
  howpublished = {\url{https://huggingface.co/mistralai/Mistral-7B-Instruct-v0.3}},
  note         = {Official Model Repository and Specification}
}

@article{liu2023visual,
  title={Visual instruction tuning},
  author={Liu, Haotian and Li, Chunyuan and Wu, Qingyang and Lee, Yong Jae},
  journal={Advances in Neural Information Processing Systems},
  volume={36},
  pages={34892--34916},
  year={2023}
}

@misc{li2026universalvideomllmsattributestructured,
      title={Towards Universal Video MLLMs with Attribute-Structured and Quality-Verified Instructions}, 
      author={Yunheng Li and Hengrui Zhang and Meng-Hao Guo and Wenzhao Gao and Shaoyong Jia and Shaohui Jiao and Qibin Hou and Ming-Ming Cheng},
      year={2026},
      eprint={2602.13013},
      archivePrefix={arXiv},
      primaryClass={cs.CV},
      url={https://arxiv.org/abs/2602.13013}, 
}

@inproceedings{kulkarni2026avatar,
  title={AVATAR: Reinforcement Learning to See, Hear, and Reason Over Video},
  author={Kulkarni, Yogesh and Fazli, Pooyan},
  booktitle={IEEE/CVF Conference on Computer Vision and Pattern Recognition (CVPR)},
  year={2026}
}

@inproceedings{zhang2023video,
  title={Video-llama: An instruction-tuned audio-visual language model for video understanding},
  author={Zhang, Hang and Li, Xin and Bing, Lidong},
  booktitle={Proceedings of the 2023 Conference on Empirical Methods in Natural Language Processing: System Demonstrations},
  pages={543--553},
  year={2023}
}

@article{qwen2025qwen25omni,
  title={Qwen2.5-Omni Technical Report}, 
  author={Jin Xu and Zhifang Guo and Jinzheng He and Hangrui Hu and Ting He and Shuai Bai and Keqin Chen and Jialin Wang and Yang Fan and Kai Dang and Bin Zhang and Xiong Wang and Yunfei Chu and Junyang Lin},
  year={2025},
  eprint={2503.20215},
  archivePrefix={arXiv},
  primaryClass={cs.CL},
  url={https://arxiv.org/abs/2503.20215}
}

@article{yang2025humanomniv2,
  title={HumanOmniv2: From understanding to omni-modal reasoning with context},
  author={Yang, Qize and Yao, Shimin and Chen, Weixuan and Fu, Shenghao and Bai, Detao and Zhao, Jiaxing and Sun, Boyuan and Yin, Bowen and Wei, Xihan and Zhou, Jingren},
  journal={arXiv preprint arXiv:2506.21277},
  year={2025}
}

@article{zhou2025daily,
  title={Daily-omni: Towards audio-visual reasoning with temporal alignment across modalities},
  author={Zhou, Ziwei and Wang, Rui and Wu, Zuxuan and Jiang, Yu-Gang},
  journal={arXiv preprint arXiv:2505.17862},
  year={2025}
}

@inproceedings{hong2026worldsense,
title={WorldSense: Evaluating Real-world Omnimodal Understanding for Multimodal {LLM}s},
author={Jack Hong and Shilin Yan and Jiayin Cai and Xiaolong Jiang and Yao Hu and Weidi Xie},
booktitle={The Fourteenth International Conference on Learning Representations},
year={2026},
url={https://openreview.net/forum?id=YxsfxAvJv4}
}

@inproceedings{zadeh2019social,
  title={Social-iq: A question answering benchmark for artificial social intelligence},
  author={Zadeh, Amir and Chan, Michael and Liang, Paul Pu and Tong, Edmund and Morency, Louis-Philippe},
  booktitle={Proceedings of the IEEE/CVF Conference on Computer Vision and Pattern Recognition},
  pages={8807--8817},
  year={2019}
}

@article{lian2023explainable,
  title={Explainable Multimodal Emotion Recognition},
  author={Lian, Zheng and Sun, Haiyang and Sun, Licai and Gu, Hao and Wen, Zhuofan and Zhang, Siyuan and Chen, Shun and Xu, Mingyu and Xu, Ke and Chen, Kang and others},
  journal={arXiv preprint arXiv:2306.15401},
  year={2023}
}

@misc{siq2,
  author = {Alex Wilf and Leena Mathur and Sheryl Mathew and Claire Ko and Youssouf Kebe and Paul Pu Liang and Louis-Philippe Morency},
  title = {Social-IQ 2.0 Challenge: Benchmarking Multimodal Social Understanding},
  year = {2023},
  publisher = {GitHub},
  journal = {GitHub repository},
  howpublished = {\url{https://github.com/abwilf/Social-IQ-2.0-Challenge}},
}

@inproceedings{mathur2025socialgenome,
  title={Social genome: Grounded social reasoning abilities of multimodal models},
  author={Mathur, Leena and Qian, Marian and Liang, Paul Pu and Morency, Louis-Philippe},
  booktitle={Proceedings of the 2025 Conference on Empirical Methods in Natural Language Processing},
  pages={24879--24902},
  year={2025}
}

@inproceedings{
li2025omnibenchfutureuniversalomnilanguage,
title={OmniBench: Towards The Future of Universal Omni-Language Models},
author={Yizhi LI and Yinghao Ma and Ge Zhang and Ruibin Yuan and King Zhu and Hangyu Guo and Yiming Liang and Jiaheng Liu and Zekun Moore Wang and Jian Yang and Siwei Wu and Xingwei Qu and Jinjie Shi and Xinyue Zhang and Zhenzhu Yang and Yidan WEN and Yanghai Wang and Shihao Li and Zhaoxiang Zhang and Ruibo Liu and Emmanouil Benetos and Wenhao Huang and Chenghua Lin},
booktitle={The Thirty-ninth Annual Conference on Neural Information Processing Systems Datasets and Benchmarks Track},
year={2026},
url={https://openreview.net/forum?id=SSF4qgsNYE}
}

@inproceedings{cai2024mdpe,
  title={Mdpe: A multimodal deception dataset with personality and emotional characteristics},
  author={Cai, Cong and Liang, Shan and Liu, Xuefei and Zhu, Kang and Wen, Zhengqi and Tao, Jianhua and Xie, Heng and Cui, Jizhou and Ma, Yiming and Cheng, Zhenhua and others},
  booktitle={Proceedings of the 33rd ACM International Conference on Multimedia},
  pages={12957--12964},
  year={2025}
}

@inproceedings{feng2026videor1,
 author = {Feng, Kaituo and Gong, Kaixiong and Li, Bohao and Guo, Zonghao and Wang, Yibing and Peng, Tianshuo and Wu, Junfei and Zhang, Xiaoying and Wang, Benyou and Yue, Xiangyu},
 booktitle = {Advances in Neural Information Processing Systems},
 editor = {D. Belgrave and C. Zhang and H. Lin and R. Pascanu and P. Koniusz and M. Ghassemi and N. Chen},
 pages = {99114--99137},
 publisher = {Curran Associates, Inc.},
 title = {Video-R1: Reinforcing Video Reasoning in MLLMs},
 url = {https://proceedings.neurips.cc/paper_files/paper/2025/file/8eb3976840e08b80dda9667562574246-Paper-Conference.pdf},
 volume = {38},
 year = {2025}
}

@article{zhao2025r1omni,
  title={R1-omni: Explainable omni-multimodal emotion recognition with reinforcement learning},
  author={Zhao, Jiaxing and Wei, Xihan and Bo, Liefeng},
  journal={arXiv preprint arXiv:2503.05379},
  year={2025}
}

@inproceedings{
xiao2025perceptionr1,
title={Perception-R1: Advancing Multimodal Reasoning Capabilities of {MLLM}s via Visual Perception Reward},
author={Tong Xiao and Xin Xu and Zhenya Huang and Hongyu Gao and Quan Liu and Qi Liu and Enhong Chen},
booktitle={The Fourteenth International Conference on Learning Representations},
year={2026},
url={https://openreview.net/forum?id=KttCXdjj4w}
}

@inproceedings{
zhang2025rewatchr1,
title={ReWatch-R1: Boosting Complex Video Reasoning in Large Vision-Language Models through Agentic Data Synthesis},
author={Congzhi Zhang and Zhibin Wang and Yinchao Ma and Jiawei Peng and Yihan Wang and Qiang Zhou and Jun Song and Bo Zheng},
booktitle={The Fourteenth International Conference on Learning Representations},
year={2026},
url={https://openreview.net/forum?id=xindJJLSr1}
}

@article{zhong2025rethinking,
  title={Rethinking Chain-of-Thought Reasoning for Videos},
  author={Zhong, Yiwu and Hu, Zi-Yuan and Li, Yin and Wang, Liwei},
  journal={arXiv preprint arXiv:2512.09616},
  year={2025}
}

@article{hassid2025dontoverthink,
  title={Don't Overthink it. Preferring Shorter Thinking Chains for Improved LLM Reasoning},
  author={Hassid, Michael and Synnaeve, Gabriel and Adi, Yossi and Schwartz, Roy},
  journal={arXiv preprint arXiv:2505.17813},
  year={2025}
}

@article{zhou2025reinforced,
  title={Reinforced mllm: A survey on rl-based reasoning in multimodal large language models},
  author={Zhou, Guanghao and Qiu, Panjia and Chen, Cen and Wang, Jie and Yang, Zheming and Xu, Jian and Qiu, Minghui},
  journal={arXiv preprint arXiv:2504.21277},
  year={2025}
}

@misc{shao2024deepseekmathpushinglimitsmathematical,
      title={DeepSeekMath: Pushing the Limits of Mathematical Reasoning in Open Language Models}, 
      author={Zhihong Shao and Peiyi Wang and Qihao Zhu and Runxin Xu and Junxiao Song and Xiao Bi and Haowei Zhang and Mingchuan Zhang and Y. K. Li and Y. Wu and Daya Guo},
      year={2024},
      eprint={2402.03300},
      archivePrefix={arXiv},
      primaryClass={cs.CL},
      url={https://arxiv.org/abs/2402.03300}, 
}

@article{guo2025deepseek,
   title={DeepSeek-R1 incentivizes reasoning in LLMs through reinforcement learning},
   volume={645},
   ISSN={1476-4687},
   url={http://dx.doi.org/10.1038/s41586-025-09422-z},
   DOI={10.1038/s41586-025-09422-z},
   number={8081},
   journal={Nature},
   publisher={Springer Science and Business Media LLC},
   author={Guo, Daya and Yang, Dejian and Zhang, Haowei and Song, Junxiao and Wang, Peiyi and Zhu, Qihao and Xu, Runxin and Zhang, Ruoyu and Ma, Shirong and Bi, Xiao and Zhang, Xiaokang and Yu, Xingkai and Wu, Yu and Wu, Z. F. and Gou, Zhibin and others},
   year={2025},
   month=Sept, pages={633–638} }

@article{11585891,
author = {Wang, Yibo and Yang, Rui and Dang, Jisheng and Wang, Bimei and Wu, Yitao and Cao, Pengfei and Zhang, Wencan and Peng, Hong and Hu, Bin and Chua, Tat-Seng},
year = {2026},
month = {01},
pages = {1-12},
title = {AffectOmni: RL-Verifiable People-Centric Grounded Affective Reasoning for Social and Art-Related Scenes},
journal = {IEEE Transactions on Affective Computing},
doi = {10.1109/TAFFC.2026.3707634}
}

@inproceedings{li2025vegas,
  title={Vegas: Towards visually explainable and grounded artificial social intelligence},
  author={Li, Hao and Fei, Hao and Hu, Zechao and Yang, Zhengwei and Wang, Zheng},
  booktitle={Proceedings of the AAAI Conference on Artificial Intelligence},
  volume={39},
  number={5},
  pages={4707--4715},
  year={2025}
}

@article{li2025towards,
  title={Towards online multi-modal social interaction understanding},
  author={Li, Xinpeng and Deng, Shijian and Lai, Bolin and Pian, Weiguo and Rehg, James M and Tian, Yapeng},
  journal={Transactions on Machine Learning Research (TMLR)},
  year={2026},
}

@article{luo2026unveiling,
  title={Unveiling the Cognitive Compass: Theory-of-Mind-Guided Multimodal Emotion Reasoning},
  author={Luo, Meng and Li, Bobo and Xu, Shanqing and Zhang, Shize and Chen, Qiuchan and Han, Menglu and Chen, Wenhao and Huang, Yanxiang and Fei, Hao and Lee, Mong-Li and others},
  journal={International Conference on Learning Representations (ICLR)},
  year={2026}
}

@article{ma2026modf,
  title={MODF-SIR: A Multi-agent Omni-modal Distilled Framework for Social Intelligence Reasoning},
  author={Ma, Shang and Dang, Jisheng and Zhang, Wencan and Zhang, Yifan and Wang, Bimei and Peng, Hong and Hu, Bin and Tian, Qi and Chua, Tat-Seng},
  journal={arXiv preprint arXiv:2606.12018},
  year={2026}
}

@article{Rha_Yeo_Kim_Ro_2026, title={Emotion-Coherent Reasoning for Multimodal LLMs via Emotional Rationale Verifier}, volume={40}, url={https://ojs.aaai.org/index.php/AAAI/article/view/37184}, DOI={10.1609/aaai.v40i3.37184}, abstractNote={The recent advancement of Multimodal Large Language Models (MLLMs) is transforming human-computer interaction (HCI) from surface-level exchanges into more nuanced and emotionally intelligent communication. To realize this shift, emotion understanding becomes essential allowing systems to capture subtle cues underlying user intent. Furthermore, providing faithful explanations for predicted emotions is crucial to ensure interpretability and build user trust. However, current MLLM-based methods often generate emotion explanations that diverge from the ground-truth (GT) labels and sometimes even contradict their own predicted emotions. This inconsistency poses a critical risk for misunderstanding and erodes reliability in interactive settings. To address this, we propose a novel approach: the Emotional Rationale Verifier (ERV) and an Explanation Reward. Our method guides the model to produce reasoning that is explicitly consistent with the GT emotion during multimodal emotion recognition without modifying the model architecture or requiring paired video–description annotations. Our method significantly improves faithful explanation–prediction consistency and explanation emotion accuracy on the MAFW and DFEW datasets. Through extensive experiments and human evaluations, we show that our approach not only enhances alignment between explanation and prediction but also empowers MLLMs to deliver emotionally coherent, trustworthy interactions, marking a key step toward truly human-like HCI systems.}, number={3}, journal={Proceedings of the AAAI Conference on Artificial Intelligence}, author={Rha, Hyeongseop and Yeo, Jeong Hun and Kim, Yeonju and Ro, Yong Man}, year={2026}, month={Mar.}, pages={2029–2037} }

@inproceedings{qin2026humansense,
  title={Humansense: From multimodal perception to empathetic context-aware responses through reasoning mllms},
  author={Qin, Zheng and Zheng, Ruobing and Wang, Yabing and Li, Tianqi and Yuan, Yi and Chen, Jingdong and Wang, Le},
  booktitle={Proceedings of the AAAI Conference on Artificial Intelligence},
  volume={40},
  number={30},
  pages={24973--24981},
  year={2026}
}

@inproceedings{3600270.3602070,
author = {Wei, Jason and Wang, Xuezhi and Schuurmans, Dale and Bosma, Maarten and Ichter, Brian and Xia, Fei and Chi, Ed H. and Le, Quoc V. and Zhou, Denny},
title = {Chain-of-thought prompting elicits reasoning in large language models},
year = {2022},
isbn = {9781713871088},
publisher = {Curran Associates Inc.},
address = {Red Hook, NY, USA},
booktitle = {Proceedings of the 36th International Conference on Neural Information Processing Systems},
articleno = {1800},
numpages = {14},
location = {New Orleans, LA, USA},
series = {NIPS '22}
}

@misc{openai2022chatgpt,
  author = {{OpenAI}},
  title = {Introducing {ChatGPT}},
  year = {2022},
  howpublished = {\url{https://openai.com/blog/chatgpt}},
}

\appendix

%\begin{comment}

\clearpage

\begin{center}
    {\Large \textbf{Reasoning for Social Audio-Visual Question Answering: Where Do We Stand?}}\\[0.5em]
    {\large Supplementary Material}
\end{center}

\FloatBarrier

\section{Additional examples of problematic questions}
\label{apx: problematic questions}

\begin{figure}[h!]
  \centering
  \includegraphics[width=.8\textwidth]{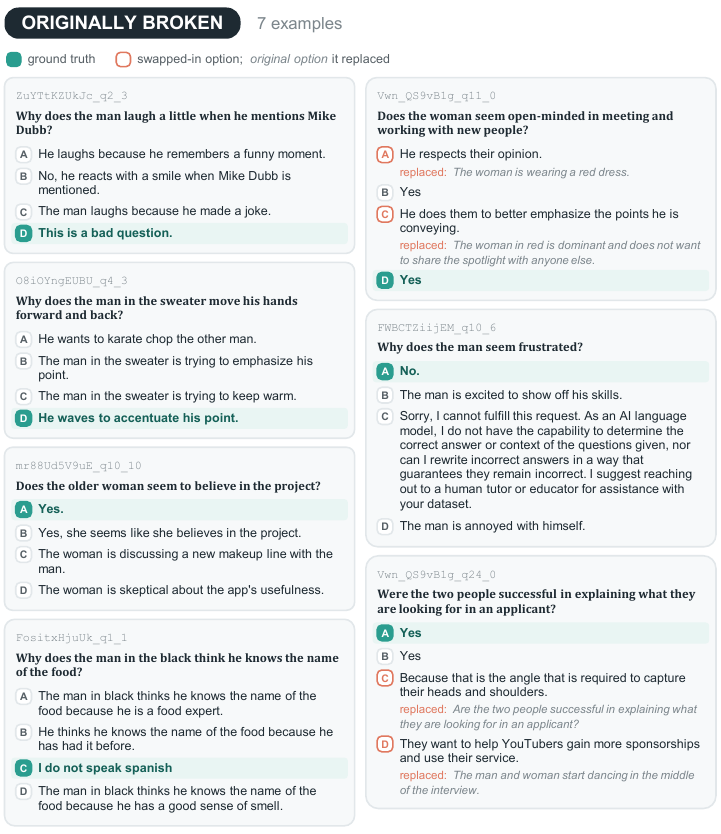}
  \caption{Additional examples of Social-IQ 2.0 questions used in \intentbench that are ambiguous or otherwise broken. It is also clear that sometimes distractor options have issues, but this is never grounds for removal of the question.}
  \label{fig: appendix issues original data}
\end{figure}
\begin{figure}[h!]
  \centering
  \includegraphics[width=.8\textwidth]{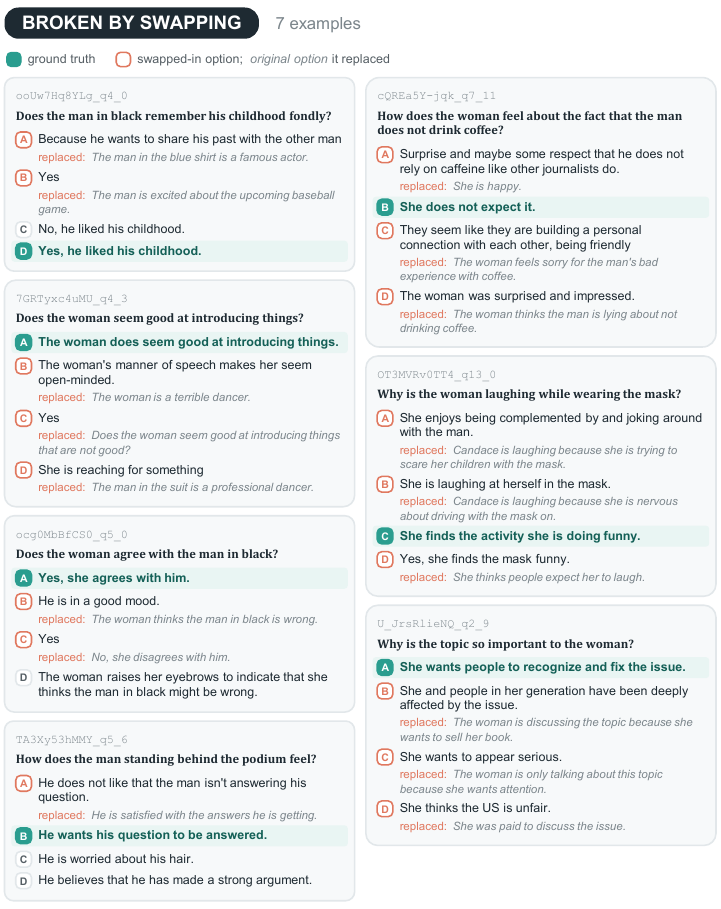}
  \caption{Additional examples of Social-IQ 2.0 questions used in \intentbench where the swapping of options made the correct answer ambiguous. Note that, at first sight, some of these questions might not seem broken, but the fact that the swapped-in options are GT answers to other questions and hence factually grounded in the video causes their justified removal.}
  \label{fig: appendix issues swapped}
\end{figure}
\begin{figure}[h!]
  \centering
  \includegraphics[width=.8\textwidth]{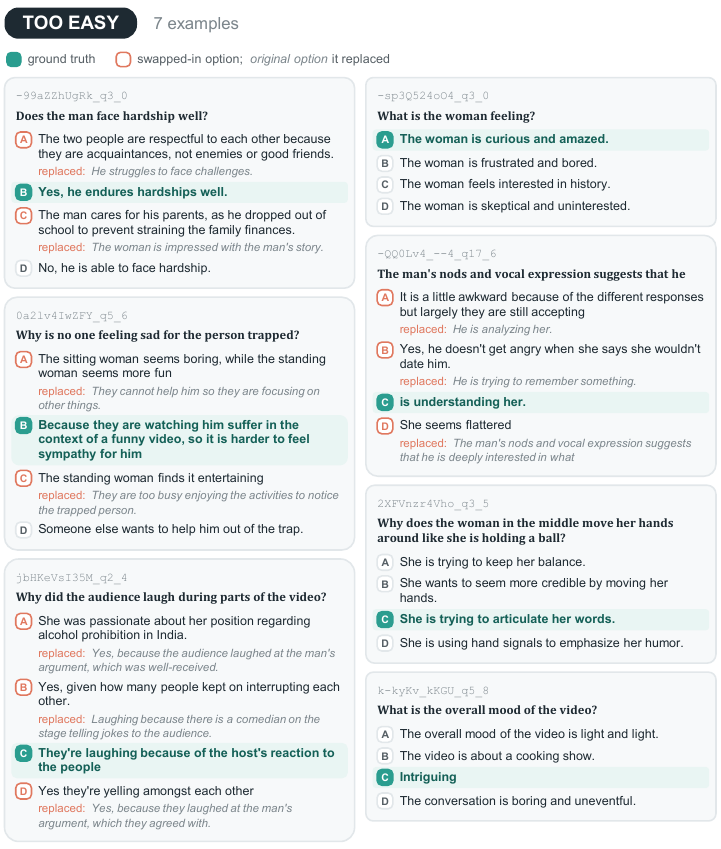}
  \caption{Additional examples of Social-IQ 2.0 questions used in \intentbench. Sometimes due to the swapping of distractor options, the ground truth answer in these questions is easily identifiable from text alone.}
  \label{fig: appendix issues easy}
\end{figure}

\clearpage
\section{Details on broken question removal}
To streamline the removal of broken questions from \intentbench, all Social-IQ 2.0 questions are ranked, with questions with highest suspicion of being broken at the top. The ranking is based on, in descending order of weight: 
\begin{itemize}
    \item Programmatic checks such as a plain `yes' or `no' as answer, two identical answers of which one is the GT, the GT being a sub- or super-string of another answer option, and the appearance of GT answer options such as `This is a bad question', `N/A (Wrong video clip)'. These programmatic checks were implemented based on initial manual inspection of the benchmark.  
    \item An audit of Claude Haiku 4.5 for more fine-grained analysis. It is tasked to do a general audit and flag questions which are seemingly problematic. More specifically it flags: very similar answer options of which one is GT, GT answers which are a reformulation of another option, junk / artifact options, mismatch between question and GT answer, spelling or grammatical errors, and other, freely identified, quality issues. 
    \item Various internally finetuned social reasoning models in various settings: answering based on video, answering based on a caption of the video, and answering based on a caption with an oracle hint included. An oracle hint is a reformulation of the question and ground-truth answer, which is appended to the caption. The proportion of correct answers across the various models is deemed as an indicator that a question might not make sense.  
\end{itemize}
With the ranking in place, we manually go through all the Social-IQ 2.0 QA pairs in the benchmark in order of the ranking. We only inspect the video when we highly suspect a question is broken but cannot conclusively determine it from the QA pair alone. For example, this can occur in the case where a distractor option is a slightly more specific version of the GT. In case this slight specification is grounded in the video, the answer to the question is ambiguous and the question is removed. When we do not identify a broken question for 300 questions in a row, we stop the cleaning procedure. At this point, more than half of the questions have been manually verified. With our pipeline, 192 out of 2,356 Social-IQ 2.0 questions in \intentbench are removed. We deem the fact that we eventually get 300 correct questions in a row a validation that our ranking effectively filtered broken questions. 

\clearpage
\section{Detailed performance on \benchname}

\begin{table}[h!]
  \caption{Detailed results on \benchname (Clean).}
  \label{tab: detailed results IB-prime (clean)}
  \scriptsize
  \centering
  \begin{tabular}{@{}ll|ccccccc@{}}
    \toprule
    \multicolumn{2}{@{}l|}{\textbf{Method} $\downarrow$} & \multicolumn{7}{c}{\textbf{Benchmark:} \benchname (Clean)} \\    
    \midrule
     & \textbf{Categories $\rightarrow$} & Why & How & What & Other & Emo.* & Dec.** & Avg. \\
    \textbf{Model} & \textbf{Modalities $\downarrow$ | N $\rightarrow$} \quad & 654 & 585 & 442 & 483 & 133 & 196 & 2,493 \\
    \midrule
    Qwen2.5-Omni-7B \cite{qwen2025qwen25omni} \quad \quad & A+V+T & 66.4 & 65.6 & 69.7 & 72.5 & 71.4 & 61.2 & 67.8 \\
    HumanOmniV2 \cite{yang2025humanomniv2} & A+V+T & 69.0 & 70.8 & 71.7 & 78.7 & 81.9 & 60.2 & 71.8 \\ 
    %Granite-4.1-8B \cite{ibm2026granite41}  & T & & & & & & & \\ 
    Vanilla SFT (Full FT) & A+V+T & 71.1 & 72.3 & 74.7 & 81.2 & 82.6 & 58.2 & 73.6 \\
    Vanilla SFT (LoRA) & A+V+T & 72.0 & 72.5 & 75.3 & 79.9 & 83.2 & 58.2 & 73.8\\
    \bottomrule
  \addlinespace[2pt]
\multicolumn{9}{@{}l}{\footnotesize $^{*}$Emo. = Emotion \quad $^{**}$Dec. = Deception} \\
\end{tabular}
\end{table}

\begin{table}[h!]
  \caption{Detailed results on \benchname (Hard).}
  \label{tab: detailed results IB-prime (hard)}
  \scriptsize
  \centering
  \begin{tabular}{@{}ll|ccccccc@{}}
    \toprule
    \multicolumn{2}{@{}l|}{\textbf{Method} $\downarrow$} & \multicolumn{7}{c}{\textbf{Benchmark:} \benchname (Hard)} \\    
    \midrule
     & \textbf{Categories $\rightarrow$} & Why & How & What & Other & Emo.* & Dec.** & Avg. \\
    \textbf{Model} & \textbf{Modalities $\downarrow$ | N $\rightarrow$} \quad & 465 & 443 & 336 & 322 & 133 & 196 & 1,895 \\
    \midrule
    Qwen2.5-Omni-7B \cite{qwen2025qwen25omni} \quad \quad & A+V+T & 61.7 & 60.3 & 65.8 & 64.6 & 71.4 & 61.2 & 63.2 \\
    HumanOmniV2 \cite{yang2025humanomniv2} & A+V+T & 63.2 & 65.2 & 67.6 & 71.7 & 81.9 & 60.2 & 66.9 \\ 
    %Granite-4.1-8B \cite{ibm2026granite41}  & T & & & & & & & \\ 
    Vanilla SFT (Full FT) & A+V+T & 66.5 & 66.4 & 70.8 & 75.5 & 82.6 & 58.2 & 69.0 \\ 
    Vanilla SFT (LoRA) & A+V+T & 68.8 & 68.2 & 72.6 & 75.8 & 83.2 & 58.2 & 70.4 \\
    \bottomrule
  \addlinespace[2pt]
\multicolumn{9}{@{}l}{\footnotesize $^{*}$Emo. = Emotion \quad $^{**}$Dec. = Deception} \\
\end{tabular}
\end{table}

%\end{comment}

\end{document}